\documentclass[preprint,12pt]{elsarticle}
\usepackage{microtype}
\usepackage{ragged2e}
\usepackage[utf8]{inputenc}
\usepackage[T1]{fontenc}
\usepackage{amsmath}
\usepackage{amsfonts}
\usepackage{amssymb}
\usepackage{enumitem}
\usepackage[version=4]{mhchem}
\usepackage{stmaryrd}
\usepackage{hyperref}
\hypersetup{colorlinks=true, linkcolor=blue, filecolor=magenta, urlcolor=cyan,}
\usepackage{graphicx}
\usepackage{subcaption}
\usepackage[export]{adjustbox}
\graphicspath{ {./images/} }
\usepackage{multirow}
\usepackage{titlesec}
\titleformat{\paragraph}
{\normalfont\normalsize\bfseries}{\theparagraph}{1em}{}
\titlespacing*{\paragraph}
{0pt}{3.25ex plus 1ex minus .2ex}{1.5ex plus .2ex}

\usepackage{makecell}
\usepackage[margin=.8in]{geometry}
\usepackage{adjustbox}
\begin{document}
\begin{frontmatter}

\title{Unified Anomaly Detection methods on Edge Device using Knowledge Distillation and Quantization }


\author[inst1]{Sushovan Jena \corref{cor1}}
\ead{sushovanjena@gmail.com}

\author[inst1]{Arya Pulkit }
\ead{aryapulkit007@gmail.com}

\author[inst1]{Kajal Singh}
\ead{kajalsinghbainsla@gmail.com}

\affiliation[inst1]{organization={School of
Computing and Electrical Engineering},
            addressline={Indian Institute of Technology}, 
            city={Mandi},
            postcode={175005}, 
            state={Himachal Pradesh},
            country={India}}

\author[inst2]{Anoushka Banerjee}
\ead{anoushka.banerjee@hitachi.co.in}

\author[inst2]{Sharad Joshi}
\ead{sharad.joshi@hitachi.co.in}

\author[inst2]{Ananth Ganesh}
\ead{ananth.ganesh@hitachi.co.in}

\cortext[cor1]{Corresponding author}
\affiliation[inst2]{organization={R\&D Center},
            addressline={ Hitachi India Pvt. Ltd }, 
            city={Bengaluru},
            postcode={560055}, 
            state={Karnataka},
            country={India}}

\author[inst1]{Dinesh Singh}
\ead{dineshsingh@iitmandi.ac.in}

\author[inst1]{Arnav Bhavsar}
\ead{arnav@iitmandi.ac.in}

\begin{abstract}
With the rapid advances in deep learning and smart manufacturing in Industry 4.0, there is an imperative for high-throughput, high-performance, and fully integrated visual inspection systems. Most anomaly detection approaches using defect detection datasets, such as MVTec AD, employ one-class models that require fitting separate models for each class. On the contrary, unified models eliminate the need for fitting separate models for each class and significantly reduce cost and memory requirements. Thus, in this work, we experiment with considering a unified multi-class setup. Our experimental study shows that multi-class models perform at par with one-class models for the standard MVTec AD dataset. Hence, this indicates that there may not be a need to learn separate object/class-wise models when the object classes are significantly different from each other, as is the case of the dataset considered.
Furthermore, we have deployed three different unified lightweight architectures on the CPU and an edge device (NVIDIA Jetson Xavier NX). We analyze the quantized multi-class anomaly detection models in terms of latency and memory requirements for deployment on the edge device while comparing quantization-aware training (QAT) and post-training quantization (PTQ) for performance at different precision widths. In addition, we explored two different methods of calibration required in post-training scenarios and show that one of them performs notably better, highlighting its importance for unsupervised tasks. Due to quantization, the performance drop in PTQ is further compensated by QAT, which yields at par performance with the original 32-bit Floating point in two of the models considered.
\end{abstract}

\end{frontmatter}

\section*{Acknowledgements}
This work is supported by Hitachi India Pvt. Ltd.

\section{Introduction}
Anomaly detection (AD), also known as outlier detection, focuses on identifying data instances that deviate significantly from the established patterns of normal behaviour. In this context, these unusual instances are referred to as anomalies, while the data points adhering to the expected patterns are considered normal \cite{pang2021,Chandola2009}. In computer vision applications, anomaly detection plays a critical role in identifying and flagging anomalous images, and one of the most promising use cases is automating the visual inspection of manufactured goods.
While supervised techniques approach the anomaly detection/segmentation problem as imbalanced binary classification or segmentation tasks, they necessitate a meticulously labelled dataset encompassing both normal and anomalous images to facilitate training. In the manufacturing industry, optical inspection tasks often lack sufficient defective samples to facilitate supervised training due to high precision standards maintained for manufacturing. Moreover, the variations in the morphology of defects are relatively ambiguous, leading to an indeterminate distribution. As a result, unsupervised or weakly supervised methods rely solely on learning from defect-free images. On the other hand, unsupervised precise segmentation of pixels, targeting regions that exhibit abnormal or novel characteristics, presents a crucial and formidable challenge in numerous computer vision domains.
There have been various works reported on the popular MVTec AD dataset \cite{bergmann2019mvtec} for unsupervised anomaly detection tasks. However, most of the existing state-of-the-art models on anomaly segmentation on MVTec AD are one-class (OC) models, where the model is trained on a particular class of object or texture and tested on the same class. This approach is way behind the current trend of multi-modal models and also incur significant cost of deployment where the model count increases with class-count. The OC models are also vulnerable to small variations inside a class as the features are highly biased towards a small domain. So, we focus on unified multi-class models which can work across large variety of objects with constraints of memory and latency. Based on performance and model size, we selected three SOTA methods, namely uninformed students (US) \cite{9157778}, reverse distillation (RD) \cite{deng2022anomaly}, and STFPM \cite{wang2021student} for 15-Class generalized training and tested the models class-wise.
As our primary goal is to deploy the models on an edge device, we explore various quantization techniques from popular frameworks such as PyTorch (Torch) and TensorRT (TRT). We have compared the performance of Torch and TRT's post-training quantization (PTQ) in 8-bit Integer (INT-8) precision in terms of performance drop and latency. In PTQ, the weights and activations are statically quantized during inference time, due to which the local minima of the converged weights with respect to the error is no more the same. This introduces a quantization error, which is responsible for a drop in performance, although with a considerable reduction in model size and latency. Here, as data calibration is a recommended part of PTQ in almost every framework, we explore two distinct ways of performing the same (training data and random normal data calibration) with marked improvement in the latter.
To compensate for the performance drop in PTQ, we also employ quantization-aware training (QAT) for fine-tuning the models, which simulated a quantization error during training, resulting in improved performance compared to post-training.

Our major contributions in this work are as follows:
\begin{enumerate}[label=(\roman*)]

\item  We experiment with generalized multi-class training of some considerably light weight methods, compare them with their one-class model performance, and suggest the generalizability of such models, which falls under a different bracket of Anomaly Segmentation methods, i.e. Unified multi-class models.

\item  The selection of the methods (Knowledge-Distillation) is strictly done from the perspective of deployment on either a CPU or an edge device and achieving real-time inference. We believe that this study would be of good significance to the community working on Unsupervised generalisation and Anomaly Detection on low-resource devices.

\item  More specifically, we provide experimental results for both memory footprints and latency as we are targeting resource-constrained environments, and we discuss that these are related not just to the network complexity but also on the Anomaly Scoring mechanism followed in respective methods.

\item  We compare the performance of the multi-class models leveraging Quantization schemes on Intel Xeon CPU and Nvidia Jetson Xavier NX in terms of AUROC and inference time, which is important for practical consideration.

\item  We analyze the PTQ performance in Torch with two different calibration strategies required for quantization, i.e., calibration using training data and random normal data, which result in a substantial gain in performance. From the deployment perspective, we demonstrate the PTQ performance with a normally distributed data calibration at different quantization precisions (16-bit Floating point (FP16), INT-8) using TRT on NVIDIA Jetson Xavier NX.

\item  We leverage QAT in Torch and compare its results with PTQ (with two calibrations) and show that the performance of QAT (INT-8) is close to that of 32-bit Floating point (FP-32).

\item  Finally, we are able to demonstrate that in some cases, even heavily quantized models do not result in a significant reduction in anomaly detection performance, which is an important practically useful revelation for this application.
\end{enumerate}

\section{Related work}
In this section, we discuss the major deep learning-based research works for anomaly detection and related approaches concerning deployment on edge devices.

\subsection{Deep learning frameworks for anomaly detection}
Some works on one-class detection include generative models like autoencoders [\cite{8953541}, \cite{bergmann2018improving}] and GANs \cite{SCHLEGL201930}. It is pertinent to highlight that these methods may sometimes yield unsatisfactory outcomes in terms of anomaly detection efficacy, largely attributed to simple per-pixel comparisons or imperfect reconstruction processes. Seminal research endeavors involving memory modules include MemSeg \cite{yang2023memseg}, which uses simulated abnormal samples and memory information in the training phase. Some efforts effectively manage data with a high-dimensional attribute space, such as DeepSVDD \cite{pmlr-v80-ruff18a} and PatchSVDD \cite{yi2020patch}.

Although some recent models show promising results on multi-class anomaly detection, they either perform less in terms of AUROC or the network architecture is far more complex and computationally expensive, which does not make them suitable for edge device deployment and achieve considerable latency [\cite{you2022unified},\cite{yao2023one},\cite{yin2023lafite}].

UniAD, \cite{you2022unified} achieves AUROC of 96.5 on multi-class paradigm compared with multi-class experiments on existing OC models. The network consists of a neighbor masked encoder consisting of masked attention and fully connected layers and a layer-wise query decoder with a feature jittering strategy. Even if its performance is better than our best performing 15-class models by $1.5-3 \%$, it is more computationally expensive due to attention \cite{vaswani2023attention} layers. Another method, One-for-all \cite{yao2023one} shows performance of 0.95 , which is very close to the 15-class model of RD. However, its architecture has vision transformers (ViT) as encoder and decoder, with proposal masking and coreset subsampling. Presence of transformer and coreset would have significantly high inference time than the models that we have experimented, they explored generative-based approach and used latent diffusion model \cite{yin2023lafite} with feature editing for reconstruction. They have also shown results on multi-class and achieved mean-auroc of 98.5. Here, the use of U-net \cite{9330594} with a diffusion model, which makes the approach costly on latency.

\subsection{Approaches involving edge device deployment}
To the best of our knowledge, we did not find existing works related to deployment of unsupervised anomaly detection models (trained on MVTec AD dataset) on edge devices. However, if we consider some other edge-device deployment cases \cite{song2021efficientdet}, which shows results of fabric defect detection on very efficient architectures like SSD and EfficientNet on Jetson TX2, where most of the models perform lower or close to our results. However, an important distinction in this case is that the datasets and models used are for supervised setting, unlike those considered in this work. Another work shows performance analysis of YOLOv3 on Jetson Xavier NX using Torch, TRT, and TensorFlow frameworks \cite{9604449}. Nonetheless, there was no comparison of quantization performance between frameworks and the results are only on one precision of quantization, i.e. FP-16 on only one model. Pioneering investigations like a study conducted by Krishnamoorthi \cite{krishnamoorthi2018quantizing} present an overview of techniques for quantizing different CNN architectures like MobileNets and ResNets (across versions) with integer weights and activations, including post-training and quantization-aware training approaches in TensorFlow. They benchmarked latencies of quantized networks on CPUs and Qualcomm DSPs,
contrary to our examination focused on unsupervised methods (especially knowledge distillation) on anomaly detection.

\section{Methodology}
We shortlist three unsupervised anomaly detection approaches based on their performance, model sizes, and deployability. The goal is to analyze the generalization behaviour of the models and their deployment using two quantization techniques, i.e., PTQ and QAT. The discussion is brief and the reader is encouraged to refer the original papers and the implementation references for more details.

\subsection{Uninformed Students}
Bergmann et al. \cite{9157778} proposes a student-teacher framework, for pixel-precise anomaly segmentation. The Knowledge Distillation first happens from a larger network like ResNet to a smaller network, a 5-layer convolutional neural network (CNN), which is the teacher. The student networks are then trained to regress upon the teacher's output as a target on the MVTec-AD dataset and so the knowledge gets distilled from teacher to students. In this process, the teacher and students' embeddings gets very close in the embedding space for normal (or non-anomalous) pixels. The anomaly score is the error between the mean predictions of the students' ensemble and the teacher's prediction. The intuition behind the anomaly score is that within anomalous regions during inference, the students' networks are expected to significantly differ from the teacher's output due to the absence of corresponding descriptors during training. This indicates the failure of student networks to generalize outside the non-anomalous data distribution. The score also considers the predictive variance of the Gaussian mixture of students' from their mean.

\subsection{Anomaly detection via reverse distillation}
\label{Anomaly detection via reverse distillation}
Reverse distillation (RD) involves passing input through the teacher (encoder) network, a bottleneck network, and then through the student (decoder) network.

The teacher (encoder) is responsible for extracting highlevel features from the input image. The bottleneck network plays a role in connecting the encoded features from the teacher network to the student network's decoder. The decoder processes the encoded features and aims to reconstruct the input image. So, the Knowledge Distillation here, happens from the Encoder to Decoder by matching the intermediate feature maps of both networks. But as the distillation happens from an encoder to decoder in the process of reconstruction of the inputs, so its termed as reverse distillation. Anomalies are detected based on the deviations of the reconstructed output from the student and the input image. Cosine similarity is used as the knowledge distillation (KD) loss for transferring knowledge between the teacher and student networks across multiple scales and layers.

\subsection{Student-Teacher Feature Pyramid Matching (STFPM)}
\label{Student-Teacher Feature Pyramid Matching (STFPM)}
Following US \cite{9157778} method, this method is an improvised framework where the multi-scale feature matching strategy is integrated to enhance anomaly detection performance. Here, the Knowledge Distillation happens from a pretrained ResNet-18 Teacher to a student ResNet-18 as we train the student to match the feature maps of Teacher network on MVTec-AD. The enhancement involves introducing hierarchical feature matching, which enables the student network to receive knowledge from multiple levels of the feature pyramid. Unlike the method of US, instead of distilling knowledge at multiple levels, the distillation happens only once, and the T-S networks are larger, i.e., ResNet-18. The strategy is to integrate both low-level and highlevel features in a complementary way to enhance anomaly detection at various sizes of anomalies.

\subsection{Quantization}
We now discuss the two quantization paradigms that we incorporated in this work, which contribute towards the practical deployment of the models on the edge device and towards model compression.

\subsubsection{Post-Training Quantization (PTQ) and Calibration}
\label{Post-Training Quantization (PTQ) and Calibration}
In PTQ, weights, and activations are quantized to INT8 from FP-32. It follows a calibration process requiring representative input data to collect statistics for each activation tensor. It records the running histogram of tensor values and $\mathrm{min} / \mathrm{max}$ values. Then, it searches the distribution in the histogram for optimal min/max values and scale factor, which would be used to perform quantization.

The search for the $\mathrm{min} / \mathrm{max}$ values and scale factor ensures the minimization of the quantization error with respect to the floating-point model. The data used for calibration should represent the range of values that the model would encounter during training or test phase. In an unsupervised setting, the test data contains very different images than the data used to train, and so it is difficult for the model to get a good scale during calibration. Hence, a random normal distribution is an optimal way to capture a generalized variance and, hence, the scale.

The quantization itself is a process that maps a floating-point value $x \in[\alpha, \beta]$ to a b-bit integer $x_{q} \in\left[\alpha_{q}, \beta_{q}\right]$,

as $x_{q}=$ round $((1 / \mathrm{s}) \cdot \mathrm{x}+\mathrm{z})$, where $\mathrm{s}$ is the scale-factor and $\mathrm{z}$ is the zero-point.

More details about quantization can be found at \cite{leimao}, with specifics for TRT and Torch at \cite{nvidia} and \cite{pytorch} respectively.




\subsubsection{Quantization-Aware Training (QAT)}
\label{Quantization-Aware Training (QAT)}
QAT enables the model to finetune and achieve better quantization-aware weights, which when quantized, should
try to preserve original performance. The framework introduces fake-quantization modules in the model architecture, i.e., quantization and dequantization modules, at the places where quantization happens during the floating-point model to quantized integer model conversion to simulate the effects of clamping and rounding brought by integer quantization. The fake-quantization modules will also monitor scales and zero points of the weights and activations. Once the QAT is finished, the floating-point model could be converted to a quantized integer model immediately using the information stored in the fake-quantization modules. During training, the rounding error keeps accumulating across samples, and as the overall loss is minimized, the rounding error also gets minimized. As a result, we have weights corresponding to the minima, which, when quantized, typically preserves the performance of the model. Thus, as the weight updating process simulates the quantization error, they converge to the minima, close to that in the floating-point case.

\section{Experimental results and analysis}
Here, we discuss the various experiments and results. First, considering our requirement of a unified multi-class model for all classes, we trained the three shortlisted methods with combined data of all classes to assess their generalization capabilities. We indicate that the models trained on a particular class and then tested only on that class (as is done in the existing works) as one-class (OC) models. Hence, we have two different models, i.e., multi-class (15-Class) and OC for each of the three methods, i.e., US \cite{9157778}, RD \cite{deng2022anomaly}, and STFPM \cite{wang2021student} (Section \ref{Comparison of One-Class model and Multi-Class model (only on FP-32)}).

Secondly, for the case of deployment on Nvidia Jetson Xavier NX (Jetson), we assessed the performance and latency of the non-quantized (FP-32) models on CPU and Jetson, which can give us a practical understanding of the speed-up in the Jetson device (Section \ref{Comparative analysis of 15-class/multi-class FP-32 models on CPU and Jetson}).

Third, to achieve better latency and lesser model size using PTQ and the deployment on the Jetson device, we considered two well established frameworks, i.e., PyTorch (Torch) and TensorRT (TRT). As part of PTQ, we explored two modes of post-training calibration (Section \ref{Performance of Post-Training Quantization (PTQ) on PyTorch with different calibration strategies}). Adhering to the best calibration method, we worked with FP-16 and INT-8 quantization on TRT (Section \ref{Performance comparison of different Quantization precisions using TensorRT on Nvidia Jetson NX}).

Finally, we note that the performance of the INT-8 quantization especially drops for PTQ. To overcome this, we then further use quantization-aware training (QAT), and demonstrate the significant improvements of (QAT) over (PTQ) (Section \ref{Difference in PTQ of PyTorch and TensorRT}).

\subsection{ Experimental Settings}
\subsubsection{ Nvidia Jetson Xavier NX and Intel Xeon CPU}
Jetson Xavier NX is an edge-computing platform from NVIDIA designed for autonomous machines and intelligent edge devices. It is built around the Xavier SoC (system-on-chip), which combines a high-performance CPU, GPU, and dedicated AI acceleration engines into a single chip. The device is built on a 6-core NVIDIA Carmel Arm 64-bit CPU and 384-core NVIDIA Volta GPU microarchitecture. This type is an advanced-level model of the Jetson family; the Xavier NX delivers a peak performance of
21TOPs. Our experiments utilized the 16GB RAM and $30 \mathrm{~W}$ power mode variant.

The CPU results are on Intel ${ }^{\circledR}$ Xeon ${ }^{\circledR} \mathrm{W}-2265,3.56$ $\mathrm{GHz}$ base frequency, built with 12 cores 2 threads per core. It is equipped with 64 GB DDR4 2933 RAM.

\subsubsection{Multi-class (or 15-class) Training}
We followed the official implementation for RD at \cite{RD} and for STFPM at \cite{STFPM}. For US, we consider the implementation at \cite{US}.
For the 15-class training of the mentioned models, we pass data of all classes in batches after shuffling to avoid bias or catastrophic forgetting. For US, we train the teacher on 15 classes. The batch size and hyper-parameter settings for each method is mentioned in Table \ref{table:hyperparameter}. All the implementations are in Torch.
\begin{table}[h]
\caption{Hyper-parameters of all methods}\label{table:hyperparameter}
\centering
\begin{tabular}{|c|l|}
\hline
Method & \multicolumn{1}{|c|}{Settings} \\
\hline
\begin{tabular}{c}
Uninformed \\
Students \\
\end{tabular} & \begin{tabular}{l}
Batch size $=1$ \\
Epochs $=150$ \\
Learning rate $=10^{-4}$ \\
Weight decay $=10^{-5}$ \\
Image size $=256 \times 256$ pixels \\
Optimizer: ADAM. \\
\end{tabular} \\
\hline
\begin{tabular}{c}
Reverse Distillation \\
\end{tabular} & \begin{tabular}{l}
Batch size $=16$ \\
Epochs $=200$ \\
Learning rate $=0.005$ \\
Weight decay $=10^{-5}$ \\
Image size $=256 \times 256$ pixels \\
Optimizer: ADAM with $\beta 1$ (exponential \\
decay for first moment estimates) $=0.5$ \\
and $\beta$ 2 (exponential decay for second \\
moment estimates) $=0.999$ \\
\end{tabular} \\
\hline
STFPM & \begin{tabular}{l}
Batch size $=32$ \\
Epochs $=398$ \\
Learning rate $=0.4$ \\
Weight decay $=10^{-4}$ \\
Momentum $=0.9$ \\
Image size $=256 \times 256$ pixels \\
Optimizer: Stochastic Gradient Descent \\
\end{tabular} \\
\hline
\end{tabular}
\end{table}

\subsubsection{ Quantization Implementation}
For PTQ and QAT of all the models, we are only quantizing the student network using default settings of Torch quantization on FBGEMM (Facebook GEneral Matrix Multiplication) backend while the teacher part of the network remains in FP-32. It is because for RD and STFPM, the teacher uses pretrained weights and only student gets trained, so quantizing only the trainable part allows us to implement QAT on that and it is also evident that this design resulted in $37 \%$ to $61 \%$ reduction in model size (across all models) and hence latency. For US, we quantized all the three student networks. Similarly, for STFPM, only the student network was quantized. In case of RD, we quantize the bottleneck and decoder (student) networks for the same, but during implementation we found that "torch.nn.ConvTranspose2D" module used in the decoder part of RD, is not supported for quantization in FBGEMM (more details are mentioned in \cite{pytorch}). So, we kept that part of decoder in FP-32 and the rest parameters are quantized to INT-8.

\subsection{Comparison of One-Class model and Multi-Class model (only on FP-32)}
\label{Comparison of One-Class model and Multi-Class model (only on FP-32)}

\begin{table}
\caption{One-Class Model vs Multi-Class (15-class) Model}~\label{table:One-Class Model vs Multi-Class Model}
\centering
\begin{tabular}{|c|c|c|}
\hline
Methods & \begin{tabular}{c}
One-class model \\
(Mean AUROC) \\
\end{tabular} & \begin{tabular}{c}
15-class model \\
(Mean AUROC) \\
\end{tabular} \\
\hline
\begin{tabular}{c}
Uninformed Students (US) \\
\end{tabular} & 0.76 & 0.79 \\
\hline
\begin{tabular}{c}
Reverse Distillation (RD) \\
\end{tabular} & 0.98 & 0.95 \\
\hline
STFPM & 0.97 & 0.94 \\
\hline
\end{tabular}
\end{table}
Table \ref{table:One-Class Model vs Multi-Class Model} shows the performance comparison of OC and 15-class models for all the methods, averaged over all classes. Fig.\ref{fig:enter-label} shows the class-wise performance. Fig.\ref{img:bcr} also depicts some qualitative results on images, where the anomaly detection heat maps are shown. Based on this, we can note the following:
\begin{enumerate}[label=(\alph*)]

\item  Table \ref{table:One-Class Model vs Multi-Class Model} shows the generalization capability of different methods. It can also be inferred from Fig.\ref{fig:enter-label} that the classwise performance of OC and 15-Class models are nearly equal (and high) for most of the classes for RD and STFPM, with US being an exception, where the performance fluctuates among some classes. Overall, the average AUROC is very similar between the $\mathrm{OC}$ and the 15-class case.
\item  It is evident that RD and STFPM, which yield high results in the OC case, are also able to generalize very well under the multi-class setup. This can be due to the presence of a larger architecture like WideResNet-50 in RD and ResNet-18 in STFPM as compared to a 5-layer architecture in US \cite{9157778}. Interestingly, in the case of US, the generalized results are in fact somewhat better than the OC case, but the absolute AUROC values are not as high as the other two methods, and it is also not consistent across classes. Hence, the RD and STFPM results may be considered more stable and reliable for generalization.

\begin{figure}
    \centering
    \includegraphics[max width=\textwidth]{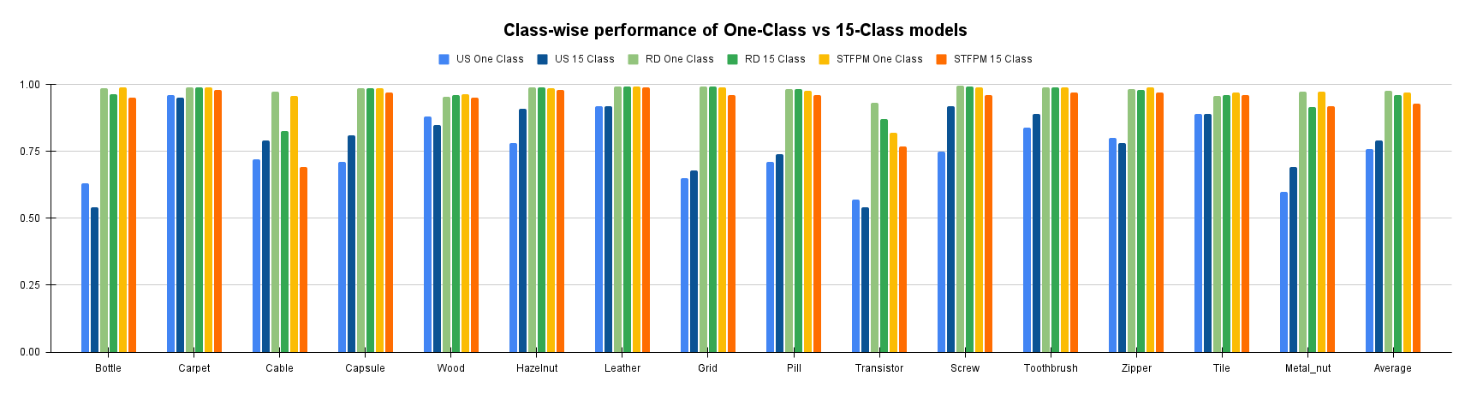}
    \caption{Graphical comparison of class-wise performance of One-class vs 15-class models of each method with dark Orange, Green, Blue line indicating 15-class or multi-class performance and light Orange, Green, Blue line indicating one-class performance.}
    \label{fig:enter-label}
\end{figure}

\begin{figure}
    \centering
    {\includegraphics[width=9cm]{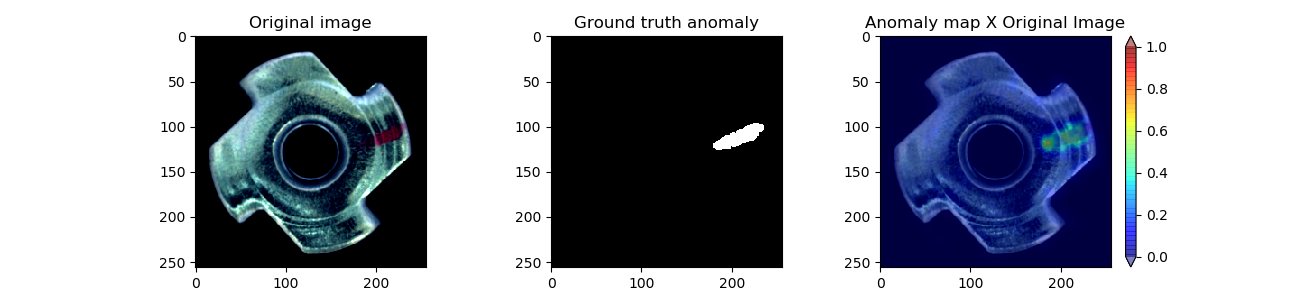} }
    \quad
    \subfloat[]{{\includegraphics[width=9cm]{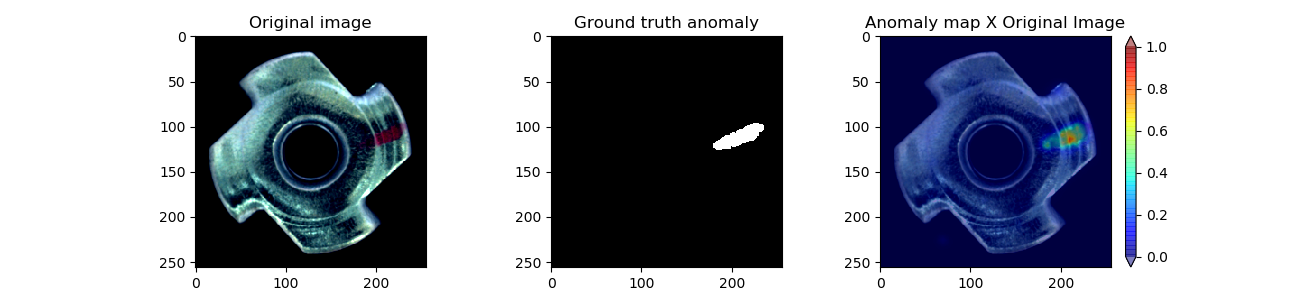} }}
    \quad
    {\includegraphics[width=9cm]{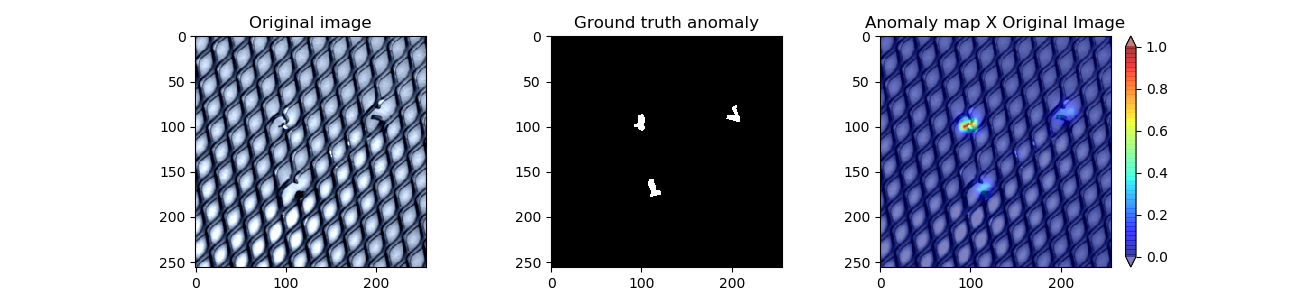} }
    \quad
    \subfloat[]{{\includegraphics[width=9cm]{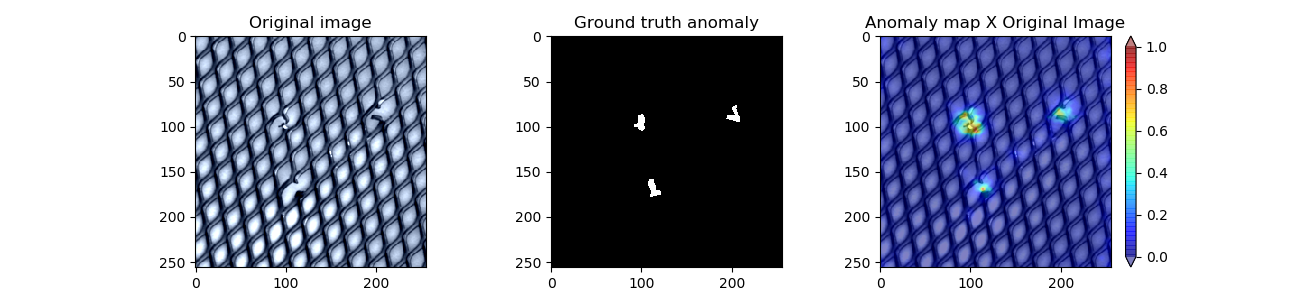} }}
    \quad
    {\includegraphics[width=9cm]{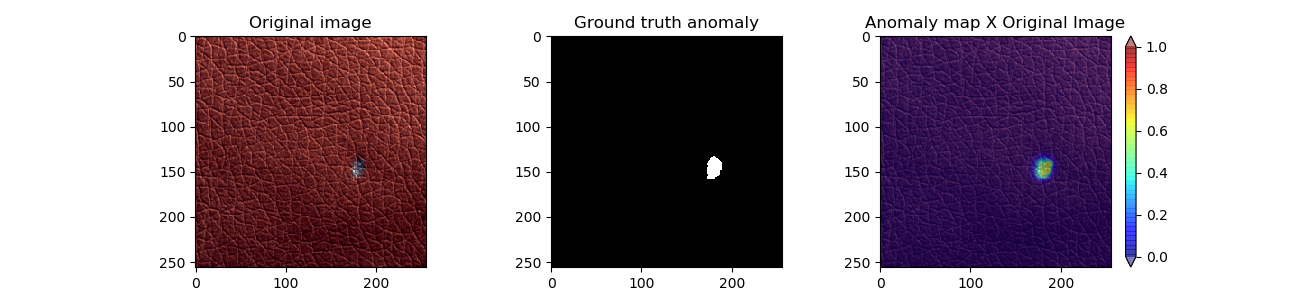}}
    \quad
    \subfloat[]{{\includegraphics[width=9cm]{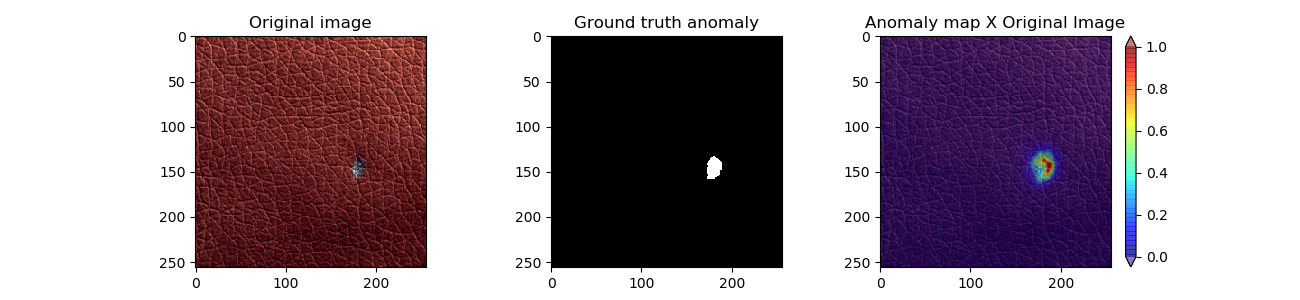} }}
    \quad
    {\includegraphics[width=9cm]{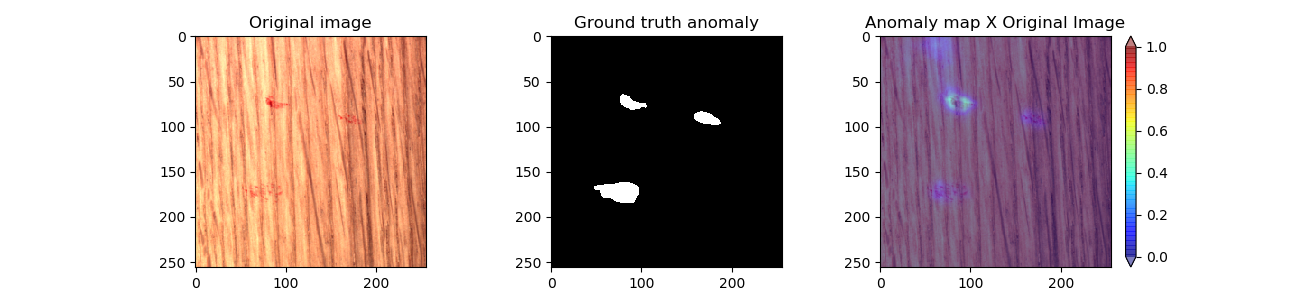}}
    \quad
    \subfloat[]{{\includegraphics[width=9cm]{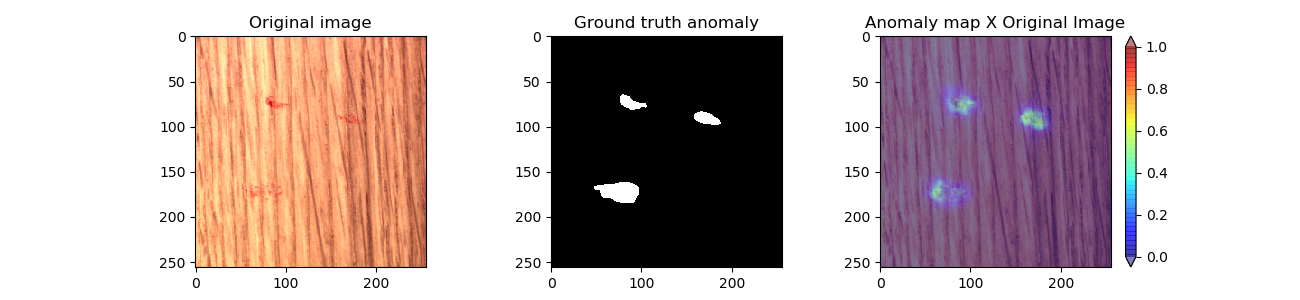} }}
    \caption{Anomaly map visualization of One-class (1st row) vs 15-Class (2nd row) results of STFPM for Object classes (a) Metal Nut (b) Grid (c) Leather (d) Wood, where the first column is the original image, 2nd column is the corresponding ground truth mask and 3rd column represents the anomaly map superimposed on the Original image.} \label{img:bcr}
\end{figure}

\item Also, the matching of intermediate feature maps during training of STFPM and a similar approach of multi-scale feature-based distillation followed in $\mathrm{RD}$, are actually able to capture the different scales of anomalies across different classes of objects/textures better. STFPM and RD approaches have leveraged combining information from different intermediate layers of the network. It is observed from Fig.\ref{fig:enter-label} that RD and STFPM show less class-wise variation in accuracy (measured in AUROC) in comparison to US, thus generalizes better across classes.

\item RD and STFPM perform very similarly, both for the $\mathrm{OC}$ as well as for the 15-class cases. However, STFPM also shows a high AUROC, with a significant improvement in latency (less inference time) than the former (Table \ref{tab: Comparison of Multi-class FP-32 models on CPU vs Jetson}). The low latency of STFPM can be attributed to its 18-layer ResNet than a 50-layer WideResNet in RD.

Also, the presence of a Bottleneck in RD, used to project the teacher model's high-dimensional representation into a low-dimensional space, to be passed to the student decoder, should also be adding more to the inference time. From Fig.\ref{img:bcr} it can be noticed that 15-class models focus on the defects with higher activation values.

\item From the qualitative perspective, it is observed in Fig.\ref{img:bcr} that the small differences in the AUROC are due to the local variation of the detected anomaly regions and not due to significant changes (e.g., false positives elsewhere). This is encouraging, as in real-world defect detection, the performance of generalized models, which are marginally lower than OC models, would not be of significant concern. This is because the lower performance is due to pixellevel errors at a local level, which are negligible, as the overall defect localization is still correct. Thus, the generalized models are able to localize the defective part as well as the OC models.

Note that in this dataset, the object appearance is quite distinct across different classes. Hence, the feature distributions of one object class are likely to be different from others. In such a case, in hindsight, it is not surprising that the anomalies, which are deviations of features from normality, will not overlap with features of other object classes, which are altogether different. This shows that in such cases, generalized models can be considered quite reliable, and there is no need for having separate models for each class, which is also validated via the experiments. Hence, in the next subsections, we only show the results for 15-class models.
\end{enumerate}

\subsection{ Comparative analysis of 15-class/multi-class FP-32 models on CPU and Jetson}
\label{Comparative analysis of 15-class/multi-class FP-32 models on CPU and Jetson}

As we proceed toward the device deployment of these methods, we now show the comparison of the Torch FP32 model between the CPU and the Jetson device in Table \ref{tab: Comparison of Multi-class FP-32 models on CPU vs Jetson}. Thus, the framework is the same (Torch) and the devices are different (CPU vs Jetson). We observe and infer the following from this:
\begin{enumerate}[label=(\alph*)]
    \item While the drop in latency is expected on the Jetson device, the order of decrease is a significant 5 to 13 times across different models. Even if we only consider the best performing models (RD and STFPM), the reduction is 5 to 7 times without any loss in AUROC. It is because of the presence of a 256-core GPU in Jetson. This comparison is intended to show real-time deployment use cases in a commonly used CPU and low-powered edge GPU.
    \item If we observe the model size and inference time across the models, an interesting observation is that even if US model is the lightest of all, it takes the highest time. This is due to the presence of a local feature extraction approach (fast dense feature extraction) \cite{bailer2018fast}, where a patch is extracted for every pixel of the whole image at once using pooling and striding layers.
    \item STFPM performs best in latency and AUROC while having the lowest model size. It has both the teacher and student as ResNet-18, where the anomaly scoring is done by taking a squared difference of the intermediate feature maps, specifically 4th, 5th and 6th layers, which have 64, 128 and 256 channels respectively. Before the squared difference, each layer is normalized across the channel dimension. This process makes the scoring process more efficient than others.
\end{enumerate}

\begin{table}[h]
\centering
\caption{Comparison of Multi-class FP-32 models on CPU vs Jetson}~\label{tab: Comparison of Multi-class FP-32 models on CPU vs Jetson}

\resizebox{\textwidth}{!}{
\begin{tabular}{|c|c|c|c|c|c|}
\hline
\multirow[t]{1}{*}{Methods} & \multirow[t]{1}{*}{}\begin{tabular}{c}
Model Size \\
(in MB) \\
\end{tabular} & \multicolumn{2}{|c|}{CPU (FP-32)} & \multicolumn{2}{|c|}{Jetson (FP-32)} \\
\cline { 3 - 6 }
 &  & \begin{tabular}{c}
Avg. Inference Time \\
(in ms) \\
\end{tabular} & \begin{tabular}{c}
Mean AUROC \\
\end{tabular} & \begin{tabular}{c}
Avg. Inference Time \\
(in ms) \\
\end{tabular} & \begin{tabular}{c}
Mean AUROC \\
\end{tabular} \\
\hline
\begin{tabular}{c}
Uninformed Students (US) \\
\end{tabular} & \begin{tabular}{c}
$26.8 (6.7+3 * 6.7)$ \\
Teacher $=6.7$ \\
Students $=3 * 6.7$ \\
\end{tabular} & 18274 & 0.79 & 1392.97 & 0.79 \\
\hline
\begin{tabular}{c}
Reverse Distillation \\
(RD) \\
\end{tabular} & \begin{tabular}{c}
$644.2 (275+269.3+$ \\
$99.9)$ \\
Teacher (Encoder) $=$ \\
275.9 \\
Bottleneck $=269.3$ \\
Decoder $=99.9$ \\
\end{tabular} & 1365.4 & 0.96 & 166.60 & 0.96 \\
\hline
STFPM & \begin{tabular}{c}
$88 (44+44)$ \\
Teacher $=44$ \\
Student $=44$ \\
\end{tabular} & 275 & 0.94 & 52.14 & 0.94 \\
\hline
\end{tabular}
}
\end{table}

As the other two methods considered, generate pixel level dimensions without going for patches, their inference is significantly accelerated. The slower performance also sheds light on the mechanism of anomaly scoring of a model having a contribution in the latency as that is different in all the three methods. Another feature adding to the time is the presence of an ensemble of three student networks along with a teacher. In US method, the anomaly scores are calculated by taking the regression error between teacher's embedding and the ensemble-mean of three students' embedding. In total, four networks (one teacher + three students) are involved during inference. It also involves a predictive variance computation where the variance of the 3 students is considered from their mean, which adds to the time.

\begin{table}
\caption{Performance of PyTorch Post-Training Quantization using Training data calibration and Random normal data calibration on CPU. Model size constitutes the FP-32 and INT-8 quantized parts of the network}~\label{table:Performance of PyTorch Post-Training Quantization}
\centering
\resizebox{\textwidth}{!}{
\begin{tabular}{|c|c|c|c|c|}
\hline
\multirow{1}{*}{Methods} & \multicolumn{3}{|c|}{PyTorch PTQ (INT-8)} & \multirow{1}{*}{}\begin{tabular}{c}
\\Model Size \\
(Teacher + Student/s) \\
(in MB)
\end{tabular} \\
\cline { 2 - 4 }
 & \begin{tabular}{c}
Avg. Inference \\
Time \\
\end{tabular} & \multicolumn{2}{|c|}{Mean AUROC} & \multirow{2}{*}{} \\
\cline { 3 - 4 }
 & (in ms) & \begin{tabular}{c}
Training Data \\ 
Calibration \\
\end{tabular} & \begin{tabular}{c}
Random Normal \\
Calibration \\
\end{tabular} &  \\
\hline
\begin{tabular}{c}
Uninformed \\
Students \\
(US) \\
\end{tabular} & 13968.88 & 0.63 & 0.64 & 
\begin{tabular}{c}
11.8\\
$\left(6.7+3^{*} 1.7\right)$ \\
\end{tabular} 
\\
\hline
\begin{tabular}{c}
Reverse \\
Distillation (RD) \\
\end{tabular} & 1018 & 0.50 & 0.75 & \begin{tabular}{c}
406.42 \\ 
$(275.9 + 67.89 + 62.63)$ \\
\end{tabular}\\
\hline
STFPM & 234.9 & 0.75 & 0.83 & \begin{tabular}{c}
55.12 \\ 
$(44 + 11.12)$ \\
\end{tabular}\\
\hline
\end{tabular}
}
\end{table}

\subsection{Performance of Post-Training Quantization (PTQ) on PyTorch with different calibration strategies}
\label{Performance of Post-Training Quantization (PTQ) on PyTorch with different calibration strategies}
To reduce the latency and memory footprint, we implemented PTQ in Torch. Typically, post-training requires a calibration process to capture the dynamic range of activations when calibrated on training data. Hence, random data calibration almost results in similar statistics.

During calibration, the scale-factor and zero-point is calculated while mapping from 32-bit to 8-bit (which is expected to reduce some performance over the FP-32 case). We have experimented with the recommended way of calibration on training data and explored another way of calibrating on a random normal distribution. Some discussions regarding this are stated below:
\begin{enumerate}[label=(\alph*)]

    \item  Although training data calibration is most common but in the case of unsupervised datasets like MVTec-AD, where the training data only consists of normal (or nonanomalous) images and test data contains both normal and anomalous images, only training data-based calibration may not consider the range of activations for anomalous images. So, we have devised another approach of calibrating on a randomly generated normal distribution, which is expected to simulate a more general subset so that the dynamic range of activations can better approximate for normal and anomalous pixels.

    \item  It can be concluded in Table \ref{table:Performance of PyTorch Post-Training Quantization}, that random normal data calibration has resulted in a significant boost in performance of $8 \%$ and $15 \%$ for STFPM and RD over calibration with training data, which is due to the above stated reason. For US, there is no improvement, where the range of activations might already have been good on training data only, which may be because of the ensemble of students already introducing some variance.
\end{enumerate}

\subsection{Performance comparison of different Quantization precisions using TensorRT on Nvidia Jetson NX}
\label{Performance comparison of different Quantization precisions using TensorRT on Nvidia Jetson NX}

We next show the results on the Jetson device but with different precisions of quantization (Table \ref{Comparison of inference metrics across methods}). Culminating from the experimentation of two calibration strategies on Torch (in Section \ref{Performance of Post-Training Quantization (PTQ) on PyTorch with different calibration strategies}), we opted for the same random normal data calibration for post-training quantization on TRT. The revelation also equips us with the computational benefit of not having to calibrate on the entire training data, which is not suitable for an edge device considering its memory and speed constraints. TRT is the recommended SDK for high performance deep learning inference on Jetson NX. We have leveraged its capabilities on the same.\\

\begin{figure}
    \centering
    \quad
    \subfloat[]{{\includegraphics[max width=100mm]{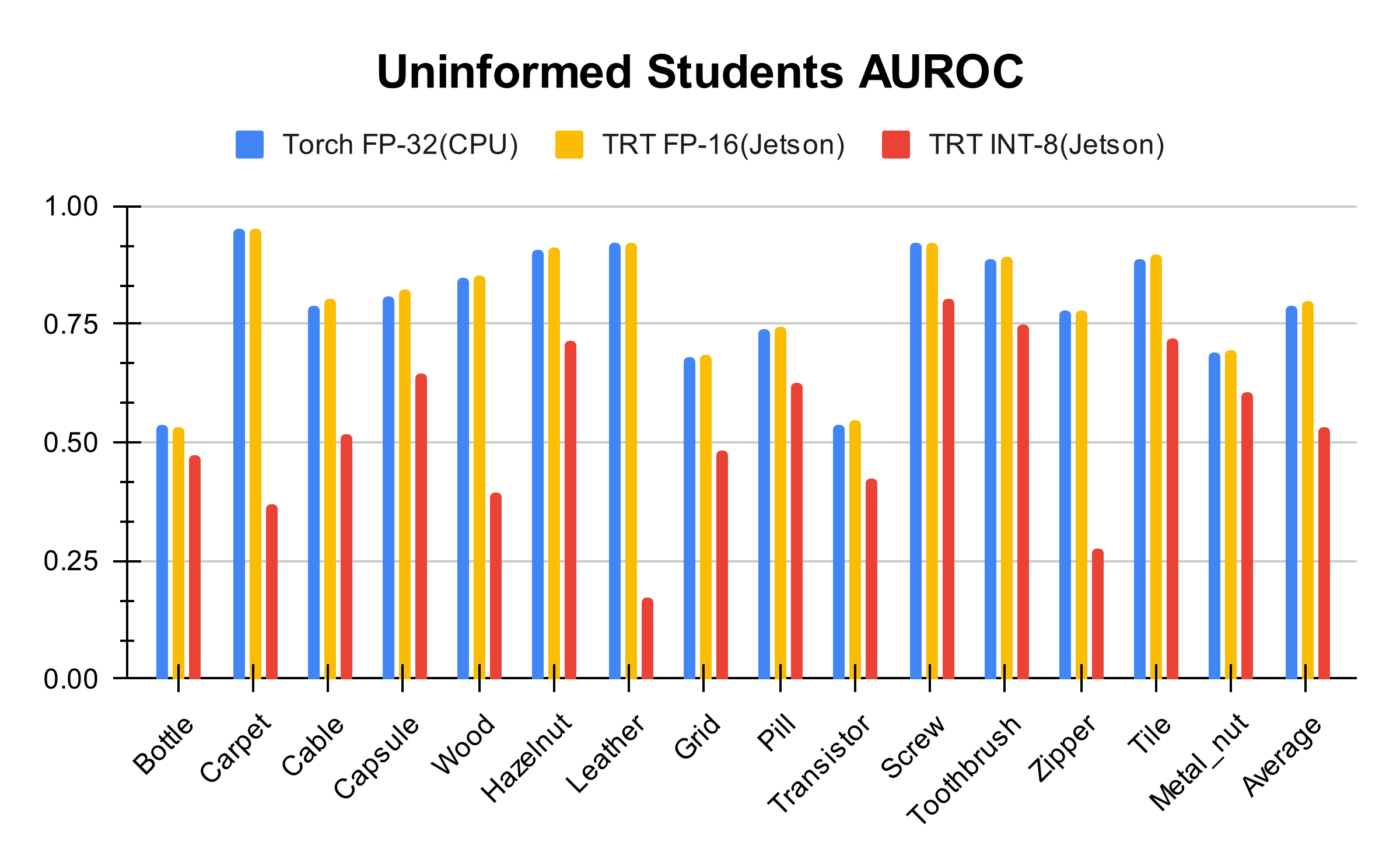} }}
    \quad
    \subfloat[]{{\includegraphics[max width=100mm]{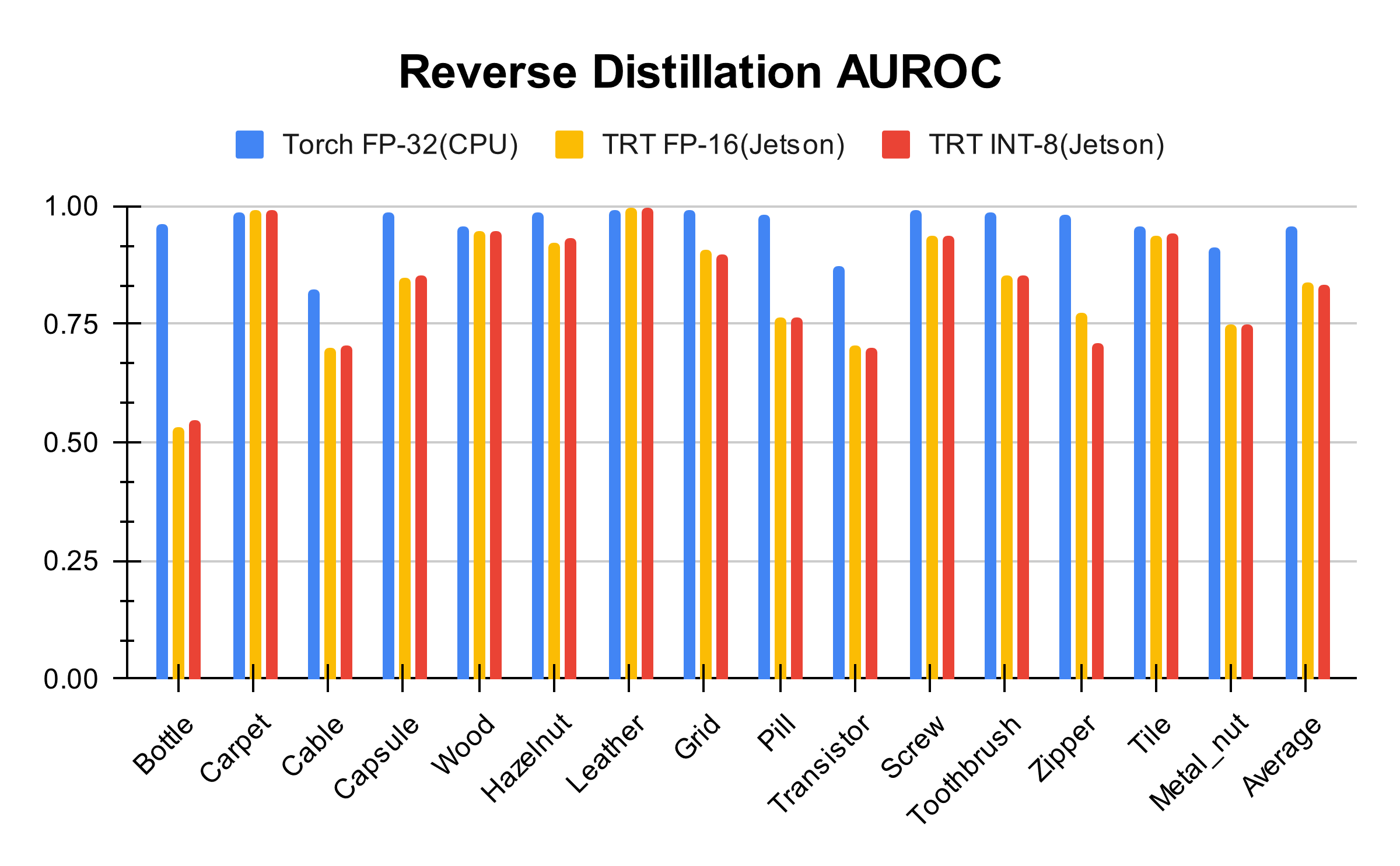} }}
    \quad
    \subfloat[]{{\includegraphics[max width=100mm]{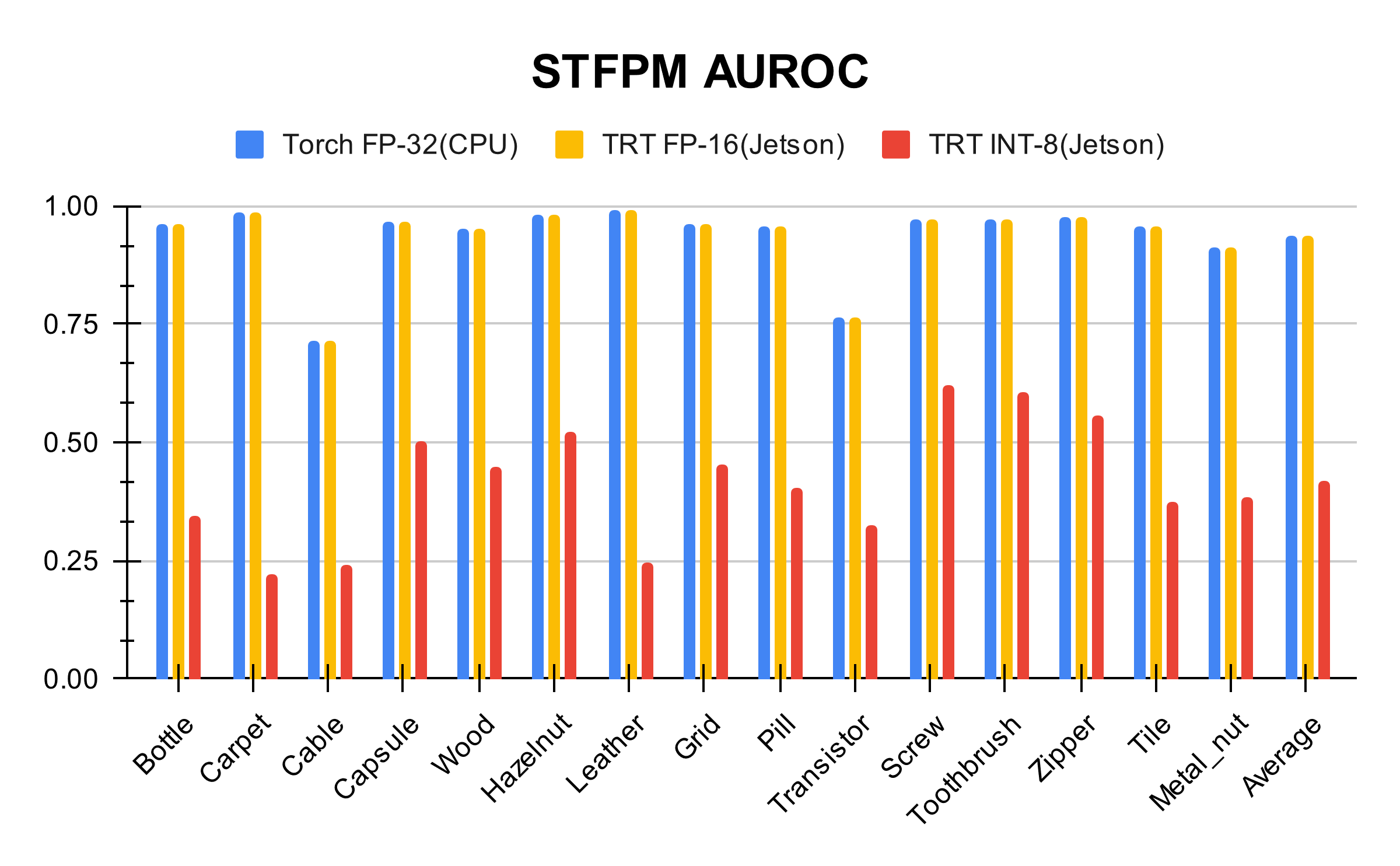} }}
    \caption{Graphical comparison of class-wise AUROC of FP-32, FP-16 and INT 8 models on Nvidia Jetson NX}
    \label{}
\end{figure}

\begin{table}[h!]
\caption{Performance of FP-32, FP-16 and INT-8 models on NVIDIA Jetson NX. The individual model sizes of teacher and student(s) are mentioned in braces. In case of TRT, for US and STFPM, it is FP-32 teacher and INT-8 quantized student. For RD, teacher (encoder) is FP-32 and bottleneck and student (decoder) are INT-8 quantized.}~\label{Comparison of inference metrics across methods}
\resizebox{\textwidth}{!}{
\centering
\begin{tabular}{|p{1.5cm}|p{1.5cm}|p{1.5cm}|p{2cm}|p{1.5cm}|p{1.5cm}|p{2cm}|p{1.5cm}|p{1.5cm}|p{2cm}|}
\hline
\multirow{2}{2cm}{Methods} & \multicolumn{3}{|c|}{PyTorch FP-32} & \multicolumn{3}{|c|}{TensorRT FP-16} & \multicolumn{3}{|c|}{TensorRT INT-8} \\ \cline{2-10}
& \centering Avg. Time (ms) & \centering AUROC & \centering Model Size (MB) & \centering Avg. Time (ms) & \centering AUROC & \centering Model Size (MB) & \centering Avg. Time (ms) & \centering AUROC & Model Size (MB) \\
\hline
 \centering US & \centering 1392.97 & \centering 0.79 & \centering 26.8 (6.7+3$\times$6.7) & \centering 699.24 & \centering 0.79 & \centering 22 (6.7+3$\times$5.1) & \centering 591.17 & \centering 0.57 & 15.4 (6.7+3$\times$2.9) \\
\hline
\centering RD & \centering 166.60 & \centering 0.96 & \centering 645.1 (275.9+ 269.3+99.9) & \centering 19.68 & \centering 0.82 & \centering 526.8 (275.9+ 176+74.9) & \centering 18.02 & \centering 0.82 & 313.64 (275.9+ 27.54+10.2) \\
\hline
STFPM & \centering 52.14 & \centering 0.94 & \centering 88 (44+44) & \centering 24.98 & \centering 0.94 & \centering 52.7 (44+8.7) & \centering 24.87 & \centering 0.92 & 48.5 (44+4.5) \\
\hline
\end{tabular}
}
\end{table}

\restoregeometry

\begin{figure}
    \centering
    {\includegraphics[width=9cm]{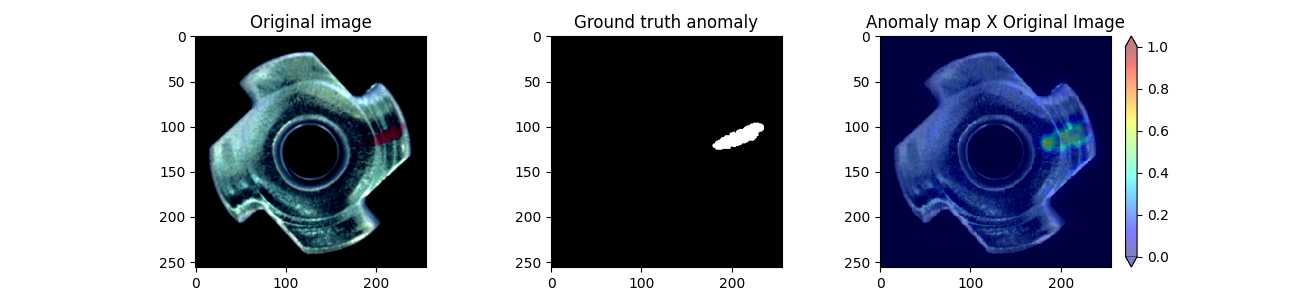} }
    \quad
    \subfloat[]{{\includegraphics[width=9cm]{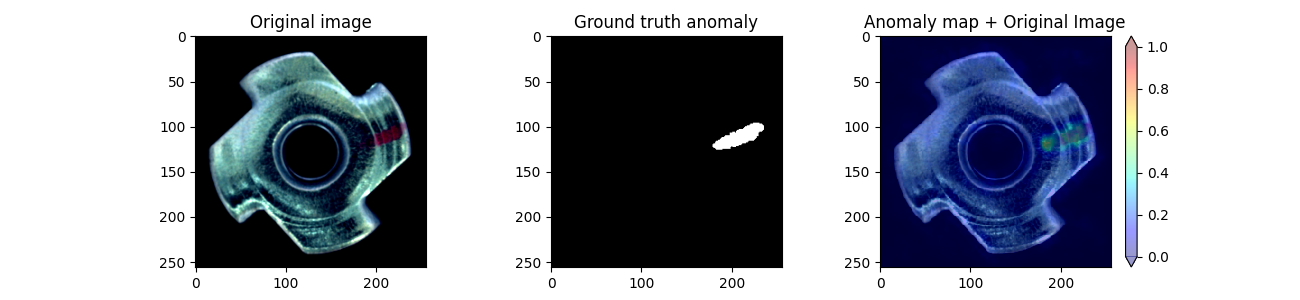} }}
    \quad
    {\includegraphics[width=9cm]{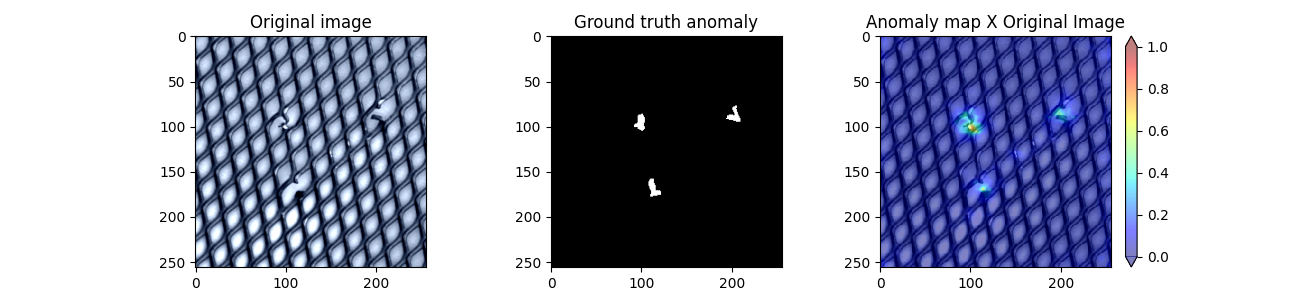} }
    \quad
    \subfloat[]{{\includegraphics[width=9cm]{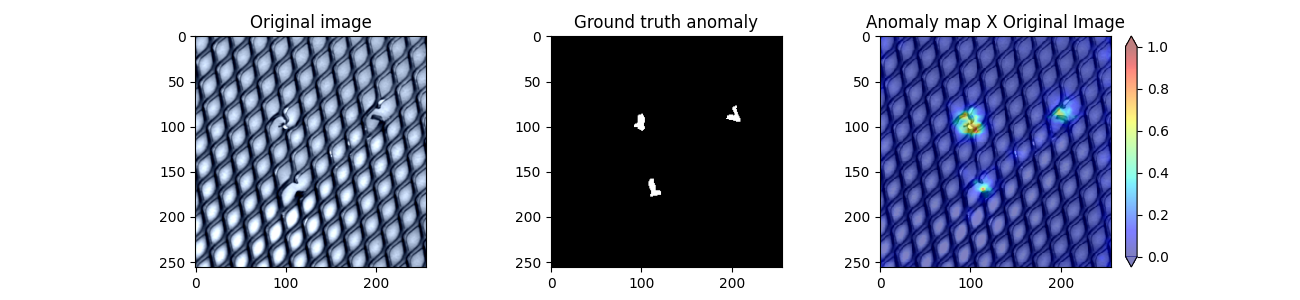} }}
    \quad
    {\includegraphics[width=9cm]{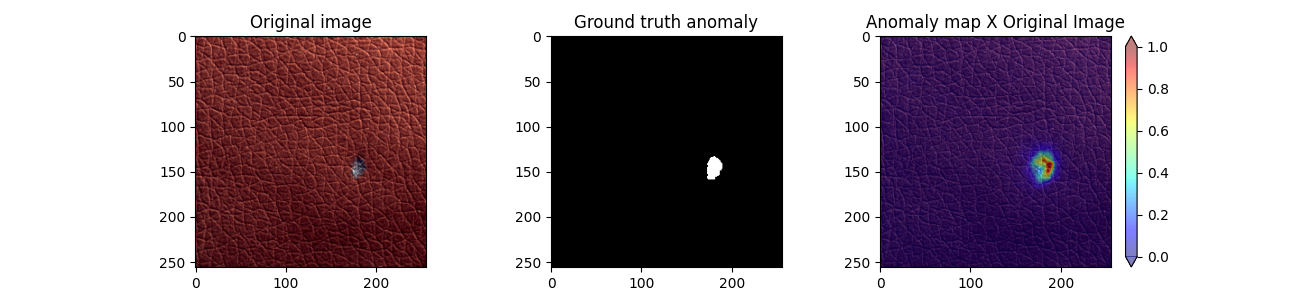}}
    \quad
    \subfloat[]{{\includegraphics[width=9cm]{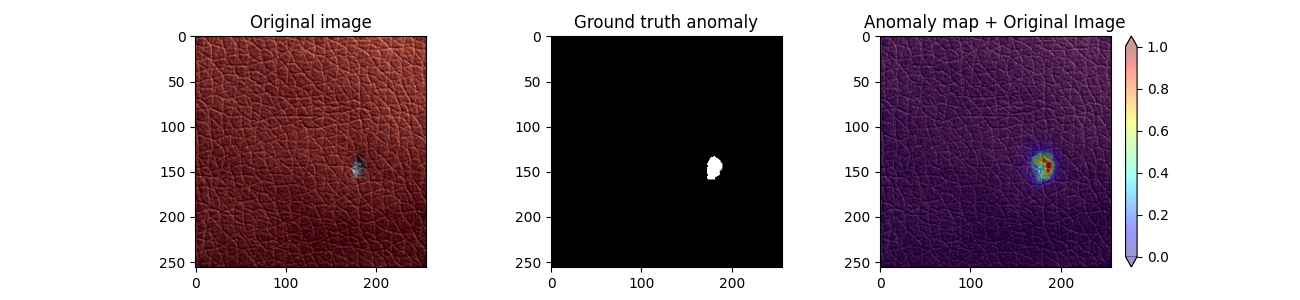} }}
    \quad
    {\includegraphics[width=9cm]{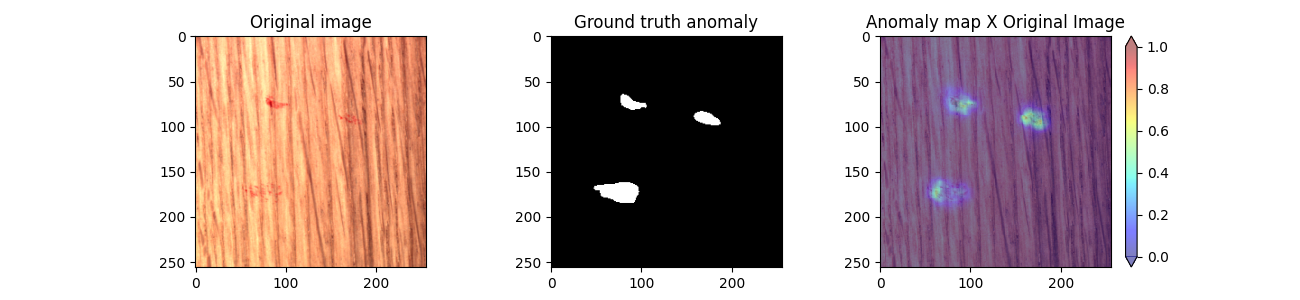}}
    \quad
    \subfloat[]{{\includegraphics[width=9cm]{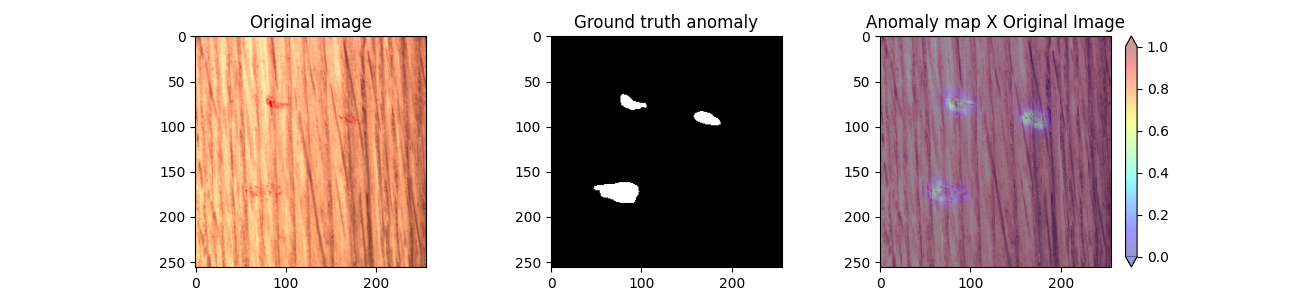} }}
    \caption{Anomaly map visualization of TRT FP-16 (1st row) vs INT-8 (2nd row) results of STFPM on NVIDIA Jetson NX, where the first column is the original image, 2nd column is the corresponding ground truth mask and 3rd column represents the Anomaly map superimposed on the Original Image.} \label{img:bcr1}
\end{figure}

The discussions on Table \ref{Comparison of inference metrics across methods} and figures are as follows: \\
\begin{enumerate}[label=(\alph*)]

\item  We note that there is a reasonably good reduction of model size for the FP-16, which further reduces for the INT8 case over the FP-32 case. As FP-16 uses half the bits compared to 32-bits for single precision, it lowers the memory usage and leads to faster inference and data-transfers. FP-16 precision is only experimented on TRT on Jetson and not on Torch as the inference time for TRT FP-32 was already 510 times lower on edge device than CPU.

\item  On the same lines, the inference time reduces significantly over the FP-32 case, especially when the FP-32 time is large (26 times and 73 times in US and RD cases), while for STFPM the FP-32 inference is itself fast, which is further increased on Jetson. However, the time difference is small between INT-8 and FP-16 versions.

\item  Despite the reduction in memory size and inference time, it is interesting to note that the mean AUROC for FP16 is not too low as compared to FP-32 model. Moreover, for the RD and especially for STFPM, even for INT-8, a high performance is maintained.

\item  As STFPM proves to be the optimal model, we consider analysing its visualizations on Jetson. Scrutinizing its anomaly maps in Fig.\ref{img:bcr1}, it is indicative that the localisation of anomalous pixels in INT-8 is almost identical to that of FP-16, which consequently signifies that the slight decrease in AUROC does not affect the comprehensive anomaly detection efficacy.

\item  For the purpose of comparison of PTQ INT8 between frameworks (Torch and TRT) between Table \ref{table:Performance of PyTorch Post-Training Quantization} and \ref{Comparison of inference metrics across methods}, Mean AUROC serves as the primary parameter and so the distinction in device (CPU or Jetson) does not affect the AUROC.
\end{enumerate}

It can be clearly observed that performance (AUROC) of RD and STFPM (the two superior models) are better in the TRT case with 0.07 to 0.09 relative difference than the Torch counterparts.

\subsection{ Difference in PTQ of PyTorch and TensorRT}
\label{Difference in PTQ of PyTorch and TensorRT}
The significant difference in AUROC performance between PTQ of Torch and TRT (both Random Normal Data calibrated) throws light on the effectiveness of the methodology followed in the two frameworks.

Below, we summarize the key differences in PTQ methodology followed in Torch vs TRT frameworks:
\begin{enumerate}[label=(\alph*)]

\item  During the process of calibration, where we capture the dynamic range of values for weights and activations of the network on a subset of training data. The values are observed in a Histogram where we get a minimum and maximum boundary. We also calculate the scale factor which is required for conversion from FP32 to INT-8. In this process, we select the optimal threshold (min. and max.) on FP32 range to map them to INT8 range. In case of TensorRT, this is done by generating many quantized distributions with different thresholds and selecting that threshold (or corresponding distribution) which minimizes the Kullback-Leibler (KL) divergence between two distributions (FP32 and INT8). As the conversion is just a reencoding of information between two models, KL-divergence (or relative entropy) measures the loss in information between the distributions. After calculation of optimal threshold and hence scale-factor, the values are quantized.\\
\item  Similar process is followed in PyTorch to calculate the min. and max. values by generating a number of quantized distributions for different $\mathrm{min} /$ max values but the error is calculated using L2 (Euclidean) Error between the FP32 distribution and quantized INT8 distribution. It involves determining the distances of each bin's content in the Histogram from the corresponding position in the two distributions. The search terminates when the optimal min/max values are found within a specified tolerance or after a maximum number of iterations.
\end{enumerate}

As the AUROC of PTQ with TensorRT is better in our experiments, this gives us an insight that minimizing the KL-divergence loss for calibration has worked better in the category of models and data considered in this study.

\subsection{ Performance analysis of QAT and PTQ}
As opposed to PTQ, which does not involve training, there is another quantization paradigm termed as quantization aware training (QAT). As QAT involves training during the quantization process, this may imply that the performance of QAT is likely to be better than PTQ. Hence, we also experiment with QAT which reveals some interesting results given in Table \ref{table:Performance of PyTorch PTQ (Random normal data calibrated) and QAT }, and discussed below:
\begin{enumerate}[label=(\alph*)]

\item It is clearly observed that the performance of QAT is significantly better than PTQ for two models. AUROC of non-quantized RD model and QAT model remains the same while for STFPM also, there is a drop of only $2 \%$.

\item In PTQ, we place observers around the weights and activations and perform a calibration process, where the training data is passed once through the model. In this process, the observers capture the dynamic range of the weights and activations, which is required to calculate the scale-factor and zero-point. Despite the calibration process, as the weights are quantized after the training, a quantization error is introduced in the model's prediction, resulting in loss of performance.

\item As discussed in Section \ref{Quantization-Aware Training (QAT)}, in QAT, we load the already trained model weights and introduce fake-quantize modules, where float values are rounded to mimic INT-8 but all computations are still done in floating-point. We then trained it for a few epochs, where the usual way of minimizing the training loss is implemented. As there is a simulated quantization error in the overall loss of the model, the same gets minimized during fine-tuning for a few epochs and we have quantize-aware weights. RD has at least four times higher latency than STFPM post QAT quantization, and only 0.04 higher AUROC point performance. Thus, STFPM can also be used where latency is critical.

\item Here, we observe that QAT clearly exhibits enhanced performance than PTQ for two methods, although the random normal data calibration method performs quite better than training data calibration. However, QAT, even for the INT-8 quantization demonstrates superior performance,
which is in fact, close to the original FP-32 performance in the case of RD and STFPM.

\item We note that for PTQ case, although the random calibration AUROC is good for RD and STFPM, there is still gap between FP-32 and quantized models, which is interestingly overcome in TRT for Jeston use case. Contrastingly, for QAT even for CPU deployment, such a gap does not exist as the top performing models (STFPM and RD) after quantization, yield results close to FP-32, obviating the need for edge device demonstration.
\end{enumerate}

\begin{table}
\caption{Performance of PyTorch PTQ (Random normal data calibrated) and QAT }\label{table:Performance of PyTorch PTQ (Random normal data calibrated) and QAT }
\centering
\resizebox{\textwidth}{!}{
\begin{tabular}{|c|c|c|c|c|c|c|}
\hline
\multirow{5}{*}{Methods} & \multicolumn{3}{|c|}{PyTorch PTQ (INT-8) in Intel Xeon CPU (Random Normal Calibrated)} & \multicolumn{3}{|c|}{PyTorch QAT (INT-8) in Intel Xeon CPU} \\
\cline{2-7}

& \begin{tabular}{c}
Avg. \\
Inference \\
Time \\
 \\
(in ms) \\
\end{tabular} & \begin{tabular}{c}
Mean \\
AUROC \\
\end{tabular} & \begin{tabular}{c}
Model Size \\
(Teacher + Student/s) \\
 \\
(in MB) \\
\end{tabular} & \begin{tabular}{c}
Avg. \\
Inference \\
Time \\
 \\
(in ms) \\
\end{tabular} & \begin{tabular}{c}
Mean \\
AUROC \\
\end{tabular} & \begin{tabular}{c}
Model Size \\
(Teacher + Student/s) \\
(in MB) \\
\end{tabular} \\
\hline
\begin{tabular}{c}
Uninformed \\
Students \\
(US) \\
\end{tabular} & 13968.88 & 0.64 & \begin{tabular}{c}
11.8 \\
$(6.7+3 * 1.7)$ \\
\end{tabular} & 14607.53 & 0.60 & \begin{tabular}{c}
11.8 \\
$(6.7+3 * 1.7)$ \\
\end{tabular} \\
\hline
\begin{tabular}{c}
Reverse Distillation \\
(RD) \\
\end{tabular} & 1018 & 0.75 & \begin{tabular}{c}
406.42 \\
$(275.9+67.89+62.63)$ \\
\end{tabular} & 1001 & 0.96 & \begin{tabular}{c}
406.42 \\
$(275.9+67.89+62.63)$ \\
\end{tabular} \\
\hline
STFPM & 234.9 & 0.83 & \begin{tabular}{c}
55.12 \\
$(44+11.12)$ \\
\end{tabular} & 235.2 & 0.92 & \begin{tabular}{c}
55.12 \\
$(44+11.12)$ \\
\end{tabular} \\
\hline
\end{tabular}
}
\end{table}

\subsection{ Overall comparative analysis of FP32, PTQ and QAT}
Finally, for a comprehensive assessment of different frameworks, precisions, we include most of the important findings from the above tables into a single one (Table \ref{Overall comparative analysis of FP-32, PTQ and QAT of STFPM method}).

Presently, PyTorch officially does not support Quantized model inference on CUDA (NVIDIA drivers). Hence, it is not possible to deploy PTQ and QAT models on NVIDIA Jetson. The same reason is behind showing performance on Intel CPU.

\begin{table}
\caption{Overall comparative analysis of FP-32, PTQ and QAT of STFPM method }\label{Overall comparative analysis of FP-32, PTQ and QAT of STFPM method}
\centering
\resizebox{\textwidth}{!}{
\begin{tabular}{|c|c|c|c|c|c|c|c|c|c|}
\hline
\multirow{6}{*}{Methods} & \multicolumn{3}{|c|}{FP-32} & \multicolumn{6}{|c|}{ PTQ (INT8) Quantized (Random Normal Calibrated) } \\
\cline {2-10}
& \multicolumn{3}{|c|}{ PyTorch (CPU, Jetson) } & \multicolumn{3}{|c|}{ PyTorch (Intel CPU) } & \multicolumn{3}{|c|}{ TensorRT (Jetson) } \\
\cline {2-10}
& \begin{tabular}{c} 
Avg. \\
Inference \\
Time \\
\\
(in ms)
\end{tabular} & \begin{tabular}{c} 
Mean \\
AUROC
\end{tabular} & \begin{tabular}{c} 
Model \\
Size \\
(Teacher + Student/s) \\
(in MB)
\end{tabular} & \begin{tabular}{c} 
Avg. \\
Inference \\
Time \\
\\
(in ms)
\end{tabular} & \begin{tabular}{c} 
Mean \\
AUROC
\end{tabular} & \begin{tabular}{c} 
Model \\
Size \\
(Teacher + Student/s) \\
(in MB)
\end{tabular} & \begin{tabular}{c} 
Avg. \\
Inference \\
Time \\
(in ms)
\end{tabular} & \begin{tabular}{c} 
Mean \\
AUROC
\end{tabular} & \begin{tabular}{c} 
Model \\
Size \\
(Teacher + Student/s) \\
(in MB)
\end{tabular} \\
\hline US & \begin{tabular}{c} 
(18274, 1392.7)
\end{tabular} & $(0.79, 0.79)$ & \begin{tabular}{c}
26.8 \\
$(6.7+3 * 6.7)$
\end{tabular} & 13968.88 & 0.64 & \begin{tabular}{c}
10.4 \\
$(6.7+3 * 1.7)$
\end{tabular} & 591.17 & 0.57 & \begin{tabular}{c}
15.4 \\
$(6.7+3 * 2.9)$
\end{tabular} \\
\hline RD & \begin{tabular}{l} 
(1365.4, 66.60)
\end{tabular} & $(0.96, 0.96)$ & \begin{tabular}{c}
645.1 \\
$(275.9 + 269.3+ 99.9)$ \\
\end{tabular} & 1018 & 0.75 & \begin{tabular}{c}
406.42 \\
$(275.9 + 67.89 + 62.63)$
\end{tabular} & 18.02 & 0.82 & \begin{tabular}{c}
313.64 \\
$(275.9 + 27.54 + 10.2)$
\end{tabular} \\
\hline STFPM & $(275, 52.14)$ & $(0.94, 0.94)$ & \begin{tabular}{c}
88 \\
$(44+44)$
\end{tabular} & 234.9 & 0.83 & \begin{tabular}{c}
55.12 \\
$(44+11.2)$
\end{tabular} & 24.87 & 0.92 & \begin{tabular}{c}
48.5 \\
$(44+4.5)$
\end{tabular} \\
\hline
 \multicolumn{10}{|c|}{} \\
\cline { 1 - 10 }
\multirow{2}{*}{Methods} & \multicolumn{9}{|c|}{PyTorch QAT (INT-8) (Intel CPU)} \\
\cline { 2 - 10 }
 & \multicolumn{3}{|c|}{
\makecell{Avg. \\Inference \\Time \\ (in ms)}}
 & \multicolumn{3}{|c|}
{\makecell{Mean \\ AUROC}} & \multicolumn{3}{|c|}
{
\makecell{Model \\Size\\ (Teacher + Student/s) \\(in MB)
}} \\
\cline {2-10}
\hline
\multirow{1}{*}{US} & \multicolumn{3}{|c|}{14607.53} & \multicolumn{3}{|c|}{0.60} &
\multicolumn{3}{|c|}{
\makecell{11.8 \\ $(6.7+3 * 1.7)$
}} \\
\hline
\multirow{1}{*}{RD} & \multicolumn{3}{|c|}{1001} & \multicolumn{3}{|c|}{0.96} & 
\multicolumn{3}{|c|}
{\makecell{406.42 \\ $(275.9+67.89+62.63)$}}
\\
\hline
\multirow{1}{*}{STFPM} & \multicolumn{3}{|c|}{235.2} & \multicolumn{3}{|c|}{0.92} &
\multicolumn{3}{|c|}
{\makecell{55.12 \\$(44+11.12)$}}\\
\hline
\end{tabular}
}
\end{table}

Finally, the overall insights from Table \ref{Overall comparative analysis of FP-32, PTQ and QAT of STFPM method} are discussed below:

\begin{enumerate}[label=(\alph*)]
\item Referring to the FP-32 column, it is a clear conclusion that an edge device such as NVIDIA Jetson is able to boost the inference speed by more than 5 times than that in CPU. This comparison is helpful in context of budget constraints in deployment of mentioned models.
\item The Avg. Inference Time and Model Size of PyTorch INT8 model is significantly lesser than that of FP-32 model on CPU with 0.11 points reduction in AUROC. This is due to the reduction in precision and hence efficient matrix computations.

\item The drop in Mean AUROC for TensorRT INT8 model on Jetson is just 0.02 as compared to FP-32 model, whereas
the drop is 0.11 in case of PyTorch INT8. Such a significant difference indicates the efficacy of PTQ methodology followed in TensorRT (discussed in Section \ref{Difference in PTQ of PyTorch and TensorRT}) over that of Pytorch.

While comparing the AUROC in Table \ref{Overall comparative analysis of FP-32, PTQ and QAT of STFPM method}, One very important consideration required is, we are not discriminating between devices such as CPU and Jetson as that does not affect the AUROC and only to be considered for inference time. We are also not considering the distinction in frameworks (Torch or TRT) as the same Torch model is converted to TRT using 'torch2trt' library and is the only possible way to deploy in Jetson as other libraries quantization is not supported (Points A and B in important issues of this section).

\item It's clearly concluded from Table \ref{Overall comparative analysis of FP-32, PTQ and QAT of STFPM method} that QAT (INT-8) performance is very close to FP32 models due to quantizeaware weights and activations resulted from finetuning, having inference time same as PTQ (INT-8) models.

\end{enumerate}

\section{Conclusion}
In this work, we focused on the task of anomaly detection on materials considering the practically important perspectives of a) generalization across object classes, b) using lightweight knowledge-distillation based models, c) further quantizing them with two schemes and analysing their performance aspects such as AUROC, latency, and model-size, and d) their deployment on an edge device. The models that we consider here also differ in their architectural designs, thus providing a variety of operational schemes, one with a patch-based knowledge distillation approach (US), other with an improved version without patching, and a multi-scale strategy (STFPM), and the last one following an encoder-decoder (RD) combined with multi-scale distillation.

First, with the experimentation on multi-class training, we establish the invariance of these to the multiclass setting for this dataset where the object appearance is quite distinct, thus obviating the need for the model-per-class paradigm. Secondly, for industrial deployment, we also assess their latency on CPU and an edge device (Nvidia Jetson $\mathrm{NX}$ ) and implement different quantization strategies to reduce the model size as well as inference time. Further, for quantization it is shown that an unconventional calibration based on the random data works much better than the standard calibration using training data, which reduces our dependence of training data. For the purpose of deployment on Jetson, we leveraged the TRT library for PTQ across two precisions, showing TRT's effectiveness over Torch for majority of models.

Finally, with an intention of further bringing the performance of the quantized model close to the un-quantized FP-32 model, both PTQ and QAT are considered, comparing their performance in CPU using Torch. This yields a very encouraging result that the quantized model with QAT (even in case of an 8-bit quantization), performs as good as the original FP-32 model for the two high performing methods.
Thus, overall, we have established that the performance of generalized, quantized models on an edge device can be as good as the original models and yet their model size and inference time can be made suitable for the operational viability in industrial settings.

\bibliographystyle{elsarticle-num} 
\bibliography{refs}

\end{document}